# Shift Aggregate Extract Networks

*Francesco Orsini*, Daniele Baracchi and Paolo Frasconi*

*Dipartimento di Ingegneria dell'Informazione, Università degli Studi di Firenze, Firenze, Italy*

We introduce an architecture based on deep hierarchical decompositions to learn effective representations of large graphs. Our framework extends classic R-decompositions used in kernel methods, enabling nested part-of-part relations. Unlike recursive neural networks, which unroll a template on input graphs directly, we unroll a neural network template over the decomposition hierarchy, allowing us to deal with the high degree variability that typically characterize social network graphs. Deep hierarchical decompositions are also amenable to domain compression, a technique that reduces both space and time complexity by exploiting symmetries. We show empirically that our approach is able to outperform current state-of-the-art graph classification methods on large social network datasets, while at the same time being competitive on small chemobiological benchmark datasets.

***Correspondence:**
*Francesco Orsini
orsini@lens.unifi.it*

**Specialty section:**
*This article was submitted to Computational Intelligence, a section of the journal Frontiers in Robotics and AI*

***Received:*** *31 October 2017*
***Accepted:*** *23 March 2018*
***Published:*** *10 April 2018*

**Citation:**
*Orsini F, Baracchi D and Frasconi P (2018) Shift Aggregate Extract Networks. Front. Robot. AI 5:42. doi: 10.3389/frobt.2018.00042*

## 1. INTRODUCTION

Structured data representations are common in application domains such as chemistry, biology, natural language, and social network analysis. In these domains, one can formulate a supervised learning problem where the input portion of the data is a graph (possibly with attributes on vertices and edges) and the output portion is a categorical or numerical label. While learning with graphs of moderate size (tens up to a few hundreds of vertices) can be afforded with many existing techniques, scaling up to large networks poses new significant challenges that still leave room for improvement, both in terms of predictive accuracy and in terms of computational efficiency.

Devising suitable representations for graph learning is crucial and nontrivial. A large body of literature exists on the subject, where graph kernels (GKs) and recurrent neural networks (RNNs) are among the most common approaches. GKs follow the classic $\mathcal{R}$-decomposition approach of Haussler (1999). Different kinds of substructures [e.g., shortest-paths (Borgwardt and Kriegel, 2005), graphlets (Shervashidze et al., 2009) or neighborhood subgraph pairs (NSPDK) (Costa and De Grave, 2010)] can be used to compute the similarity between two graphs in terms of the similarities of their respective sets of parts. RNNs (Goller and Kuechler, 1996; Sperduti and Starita, 1997; Scarselli et al., 2009) unfold a template (with shared weights) over each input graph and construct the vector representation of a node by recursively composing the representations of its neighbors. These representations are typically derived from a loss minimization procedure, where gradients are computed by the backpropagation through structure algorithm (Goller and Kuechler, 1996). Micheli (2009) proposed the architecture *neural networks for graphs* (NN4G) to learn from graph inputs with feedforward neural networks. One advantage of RNNs over GKs is that the vector representations of the input graphs are learned rather than handcrafted.

Most GK- and RNN-based approaches have been applied to relatively small graphs, such as those derived from molecules (Bianucci et al., 2000; Borgwardt and Kriegel, 2005; Ralaivola et al., 2005), natural language sentences (Socher et al., 2011) or protein structures (Baldi and Pollastri, 2003; Vullo and Frasconi, 2004; Borgwardt et al., 2005). On the other hand, large graphs (especially social networks) typically exhibit a highly-skewed degree distribution that originates a huge vocabulary of distinct subgraphs. This scenario makes finding a suitable representation much harder: kernels based on subgraph matching would suffer diagonal dominance (Schoelkopf et al., 2002), while RNNs would face the problem of composing a highly variable number of substructure representations in the recursive step. Recent work by Yanardag and Vishwanathan (2015) proposes deep graph kernels (DGK) to upgrade existing graph kernels with a feature reweighing schema that employs CBOW/Skip-gram embedding of the substructures. Another recent work by Niepert et al. (2016) casts graphs into a format suitable for learning with convolutional neural networks (CNNs). These methods have been applied successfully to small graphs but also to graphs derived from social networks.

A related but distinct branch of research focuses on the problem of predicting relations in relational structures. For example, classifications of nodes in a graph can be seen as the problem of predicting relations of arity one. Similarly, link prediction Liben-Nowell and Kleinberg (2007) can be seen as the problem of predicting relations of arity two. Methods for solving problems in this class include statistical relational learning Getoor and Taskar (2007), probabilistic inductive logic programming De Raedt et al. (2008), kernel methods (e.g., Frasconi et al., 2014), and convolutional neural networks (e.g., Atwood and Towsley, 2016; Kipf and Welling, 2017). Few of these methods are also suitable for graph classification or regression problems. Exceptions include Frasconi et al. (2014), which however does not learn representations from data, and Atwood and Towsley (2016).

In this paper, we introduce a novel architecture for learning graph representations (and therefore suitable for solving the graph classification problem), called shift-aggregate-extract network (SAEN). Structured inputs are first decomposed in a hierarchical fashion. A feedforward neural network is then *unfolded* over the hierarchical decompositions using *shift*, *aggregate*, and *extract* operations (see section 4). Finally, gradient descent learning is applied to the resulting network.

Like the flat $\mathcal{R}$-decompositions commonly used to define kernels on structured data (Haussler, 1999), $\mathcal{H}$-decompositions are based on the *part-of* relation, but allow us to introduce a deep recursive notion of *parts of parts*. At the top level of the hierarchy lies the *whole* data structure. Objects at each intermediate level are decomposed into parts that form the subsequent level of the hierarchy. The bottom level consists of atomic objects, such as individual vertices, edges, or small graphlets.

SAEN compensates some limitations of recursive neural networks by adding two synergetic degrees of flexibility. First, it unfolds a neural network over a hierarchy of parts rather than using the edge set of the input graph directly; this makes it easier to deal with very high degree vertices. Second, it imposes weight sharing and fixed size of the learned vector representations on a per level basis instead of globally; in this way, more complex parts may be embedded into higher dimensional vectors, without forcing to use excessively large representations for simpler parts.

A second contribution of this work is a *domain compression* algorithm that can significantly reduce memory usage and runtime. It leverages mathematical results from lifted linear programming (Mladenov et al., 2012) in order to exploit symmetries and perform a lossless compression of $\mathcal{H}$-decompositions.

The paper is organized as follows. In section 2 we introduce $\mathcal{H}$-decompositions, a generalization of Haussler's $\mathcal{R}$-decomposition relations (Haussler, 1999). In section 4 we describe SAEN, a neural network architecture for learning vector representations of $\mathcal{H}$-decompositions. Furthermore, in section 5 we explain how to exploit symmetries in $\mathcal{H}$-decompositions in order to reduce memory usage and runtime. In section 6 we report experimental results on several number of real-world datasets. Finally, in section 7 we discuss some related works and draw some conclusions in section 8.

## 2. $\mathcal{H}$-DECOMPOSITIONS

In this section, we define a deep hierarchical extension of Haussler's $\mathcal{R}$-decomposition relation (Haussler, 1999).

An $\mathcal{H}$-decomposition is formally defined as the triple $(\{S_l\}_{l=0}^L, \{\mathcal{R}_{l,\pi}\}_{l=1}^L, X)$ where:

- $\{S_l\}_{l=0}^L$ are disjoint sets of objects $S_l$ called levels of the hierarchy. The bottom level $S_0$ contains atomic (i.e., non-decomposable) objects, while the other levels $\{S_l\}_{l=1}^L$ contain compound objects, $s \in S_l$, whose parts $s' \in S_{l-1}$ belong to the preceding level, $S_{l-1}$.
- $\{\mathcal{R}_{l,\pi}\}_{l=1}^L$ is a set of $l, \pi$-parametrized $\mathcal{R}_{l,\pi}$-convolution relations, where $\pi \in \Pi_l$ is a membership type from a finite alphabet $\Pi_l$ of size $n(l) = |\Pi_l|$. At the bottom level, $n(0) = 1$. A pair $(s, s') \in S_l \times S_{l-1}$ belongs to $\mathcal{R}_{l,\pi}$ iff $s'$ is part of $s$ with membership type $\pi$. For notational convenience, the parts of $s$ are denoted as $\mathcal{R}_{l,\pi}^{-1}(s) = \{s' | (s', s) \in \mathcal{R}_{l,\pi}\}$.
- $X$ is a set $\{\mathbf{x}(s)\}_{s \in S_0}$ of $p$-dimensional vectors of attributes assigned to the elements $s$ the bottom layer $S_0$.

The membership type $\pi$ is used to represent the roles of the parts of an object. For $L > 1$, an $\mathcal{H}$-decomposition is a multilevel generalization of the classic $\mathcal{R}$-convolution. It represents structured data as a hierarchy of $\pi$-parametrized parts.

An example of a 4-level $\mathcal{H}$-decomposition is shown in **Figure 1** where a top-level graph in $S_3$ is decomposed into a set of $r$-neighborhood (for radius $r \in \{1, 2\}$) subgraphs $Ball \in S_2$ (see **Figure 2** for a pictorial representation of the parts) and the radius $r$ is used as the membership type. Level $S_1$ consists of edges from the $r$-neighborhood subgraphs. Finally, each edge is decomposed as pairs of vertices $V \in S_0$. The elements of the $\mathcal{R}_{l,\pi}$-convolution are pictorially shown as directed arcs. Since membership types $\pi$ for edges and vertices would be all identical their label is not represented in the picture.

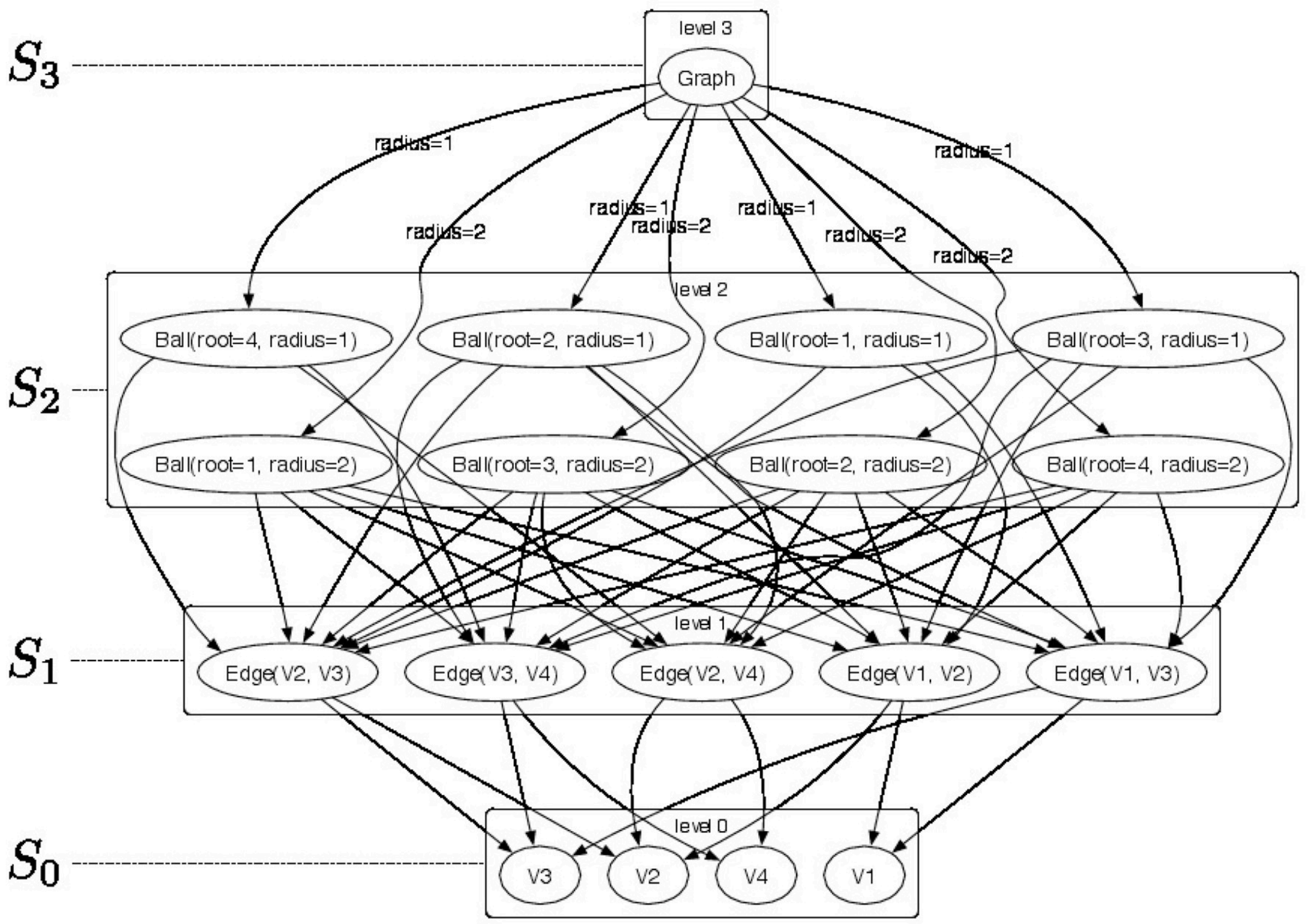

**FIGURE 1 |** Pictorial representation of a sample $\mathcal{H}$-decomposition. We produce a 4-level $\mathcal{H}$-decomposition by decomposing graph *Graph* $\in S_3$ into a set of *radius*-neighborhood (*radius* $\in \{1, 2\}$) subgraphs *Ball* $\in S_2$ and employ their *radius* as membership type. Furthermore, we extract edges *Edge* $\in S_1$ from the *radius*-neighborhood subgraphs. Finally, each edge is decomposed in vertices $V \in S_0$. The elements of the $\mathcal{R}_{l,\pi}$-convolution are pictorially shown as directed arcs. Since membership types $\pi$ for edges and vertices would be all identical their label is not represented in the picture.

Additional examples of $\mathcal{H}$-decompositions are given in the following section.

## 3. INSTANCES OF $\mathcal{H}$-DECOMPOSITIONS

We describe two $\mathcal{H}$-decompositions based on ego graphs and on nested ego graphs. They are inspired from closely related graph kernels.

*Definition 1.* The subgraph of $G = (V, E)$ induced by $V_g \subset V$ is the graph $g = (V_g, E_g)$ where $E_g = \{(u, v) \in E : u \in V_g, v \in V_g\}$.

*Definition 2.* The ego graph $g_{v,r}$ of $G = (V, E)$ with root $v \in V$ and radius $r$ is the subgraph of $G$ induced by the set of vertices whose shortest path distance from $v$ is at most $r$.

### 3.1. Ego Graph Decomposition

The ego graph $\mathcal{H}$-decomposition (EGD) has $L = 3$ levels defined as follows (see **Figure 3**):

- Level 2 consists of the whole attributed graph $G = (V, E, \mathbf{x})$ where $\mathbf{x}$ is a labeling function that attaches a $p$-dimensional vector of attributes $\mathbf{x}(v)$ to each vertex $v$.
- Level 1 consists of all ego graphs $g_{v,r}$ with roots $v \in V$ and $r \in [0, R]$. The $\pi$-type of $g_{v,r}$ is simply $r$. Note that for $r = 0$, all ego graphs $g_{v,0}$ consist of single vertices.
- Level 0 consists of single vertices with two possible $\pi$-types: ROOT and ELEM to specify whether a vertex $v$ is the root $g_{v,r}$ or not.

### 3.2. Nested Ego Graph Decomposition

The nested ego graph $\mathcal{H}$-decomposition (NEGD) has $L = 3$ levels defined as follows:

- Level 2 ($S_2$) consists of the whole attributed graph $G = (V, E, f_V, f_E)$ where $f_V$ and $f_E$ are two labeling functions that attach respectively a $p$-dimensional vector of attributes $f_V(v)$ to each vertex $v$ and a symbol $f_E(u, w)$ from a finite alphabeth $\Pi_1$ to each edge $(u, w)$.
- Level 1 ($S_1$) consists of all ego graphs $g_{v,1} = (V_v, E_v)$ with roots $v \in V$. The $\pi$-type of $g_{v,1}$ is the number of vertices $|V_v|$.
- Level 0 ($S_0$) consists of the ego graphs $g_{w,1}$, $\forall w \in V_v$, with $\pi$-type ROOT if $w = v$, or $\pi$-type $f_E(v, w)$ otherwise.
- A bijection $\mathbf{x} : S_0 \to \mathbb{N}$ associates a different identifier to each distinct ego graph in $S_0$, i.e., $\mathbf{x}(s_1) = \mathbf{x}(s_2) \iff s_1 = s_2$, $\forall s_1, s_2 \in S_0$.

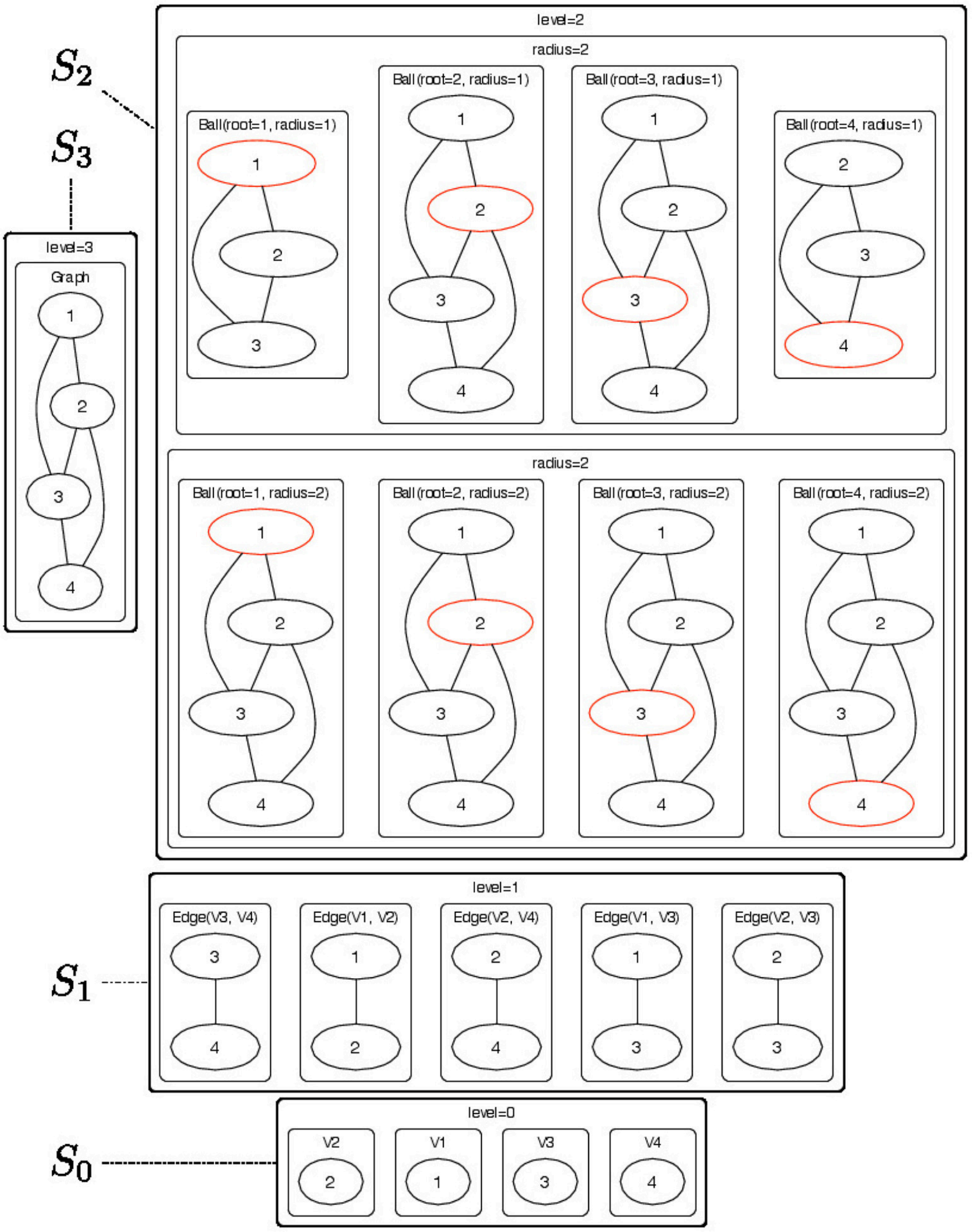

**FIGURE 2 |** Pictorial representation of the substructures that are contained in each node of the $\mathcal{H}$-decomposition showed in **Figure 1**. The objects of the $\mathcal{H}$-decomposition are grouped to according their $S_l$ sets ($l = 0, \ldots, 3$). For each *radius*-neighborhood subgraph we show the root node in red.

## 4. LEARNING REPRESENTATIONS WITH SAEN

A shift-aggregate-extract network (SAEN) is a composite function that maps objects at level $l$ of an $\mathcal{H}$-decomposition into $d(l)$-dimensional real vectors. It uses a sequence of parametrized functions $\{f_0, \ldots, f_L\}$, for example a sequence of neural networks with parameters $\theta_0, \ldots, \theta_L$ that will be trained during the learning. At each level, $l = 0, \ldots, L$, each function $f_l : \mathbb{R}^{n(l)d(l)} \to \mathbb{R}^{d(l+1)}$ operates as follows (see **Figure 4** for an illustration):

1. It receives as input the *aggregate* vector $\mathbf{a}_l(s)$ defined as:

$$\mathbf{a}_l(s) = \begin{cases} \mathbf{x}(s) & \text{if } l = 0 \\ \sum_{\pi \in \Pi_l} \sum_{s' \in \mathcal{R}^{-1}_{l,\pi}(s)} \mathbf{z}_\pi \otimes \mathbf{h}_{l-1}(s') & \text{if } l > 0 \end{cases} \quad (1)$$

   where $\mathbf{x}(s)$ is the vector of attributes for object $s$.
2. It *extracts* the vector representation of $s$ as

$$\mathbf{h}_l(s) = f_l(\mathbf{a}_l(s); \theta_l) \quad (2)$$

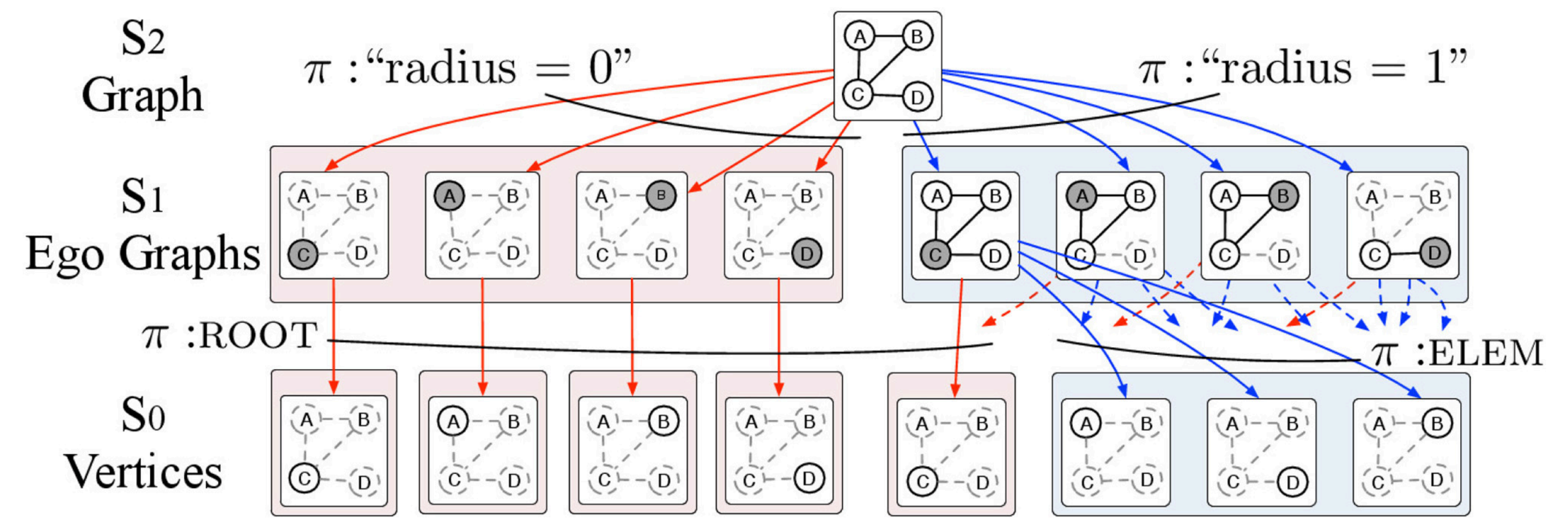

**FIGURE 3 |** The EGD is an $\mathcal{H}$-decomposition structured in 3 levels. Level 2 contains the input attributed graph $G = (V, E, X)$ where $V$ is the set of vertices and $E$ is the set of edges and $X$ is a set of $p$-dimensional vectors of attributes assigned to the vertices $v \in V$ of the graph $G$. The input graph $G$ is then decomposed into ego graphs $g$ of radius $r = 0, \ldots, R$ where $R$ is the maximum radius that we allow in the decomposition. The ego graphs $g$ are elements of level 1 and are parts of $G$ with $\pi$-type $r$. Ego graphs $g$ are further decomposed into vertices $v$. We use the $\pi$-types ROOT and ELEM to specify whether a vertex $v$ is the root of the ego graph $g$ or just an element respectively. The vertices $v$ which are the elements of level 0 and are labeled with vectors of vertex attributes.

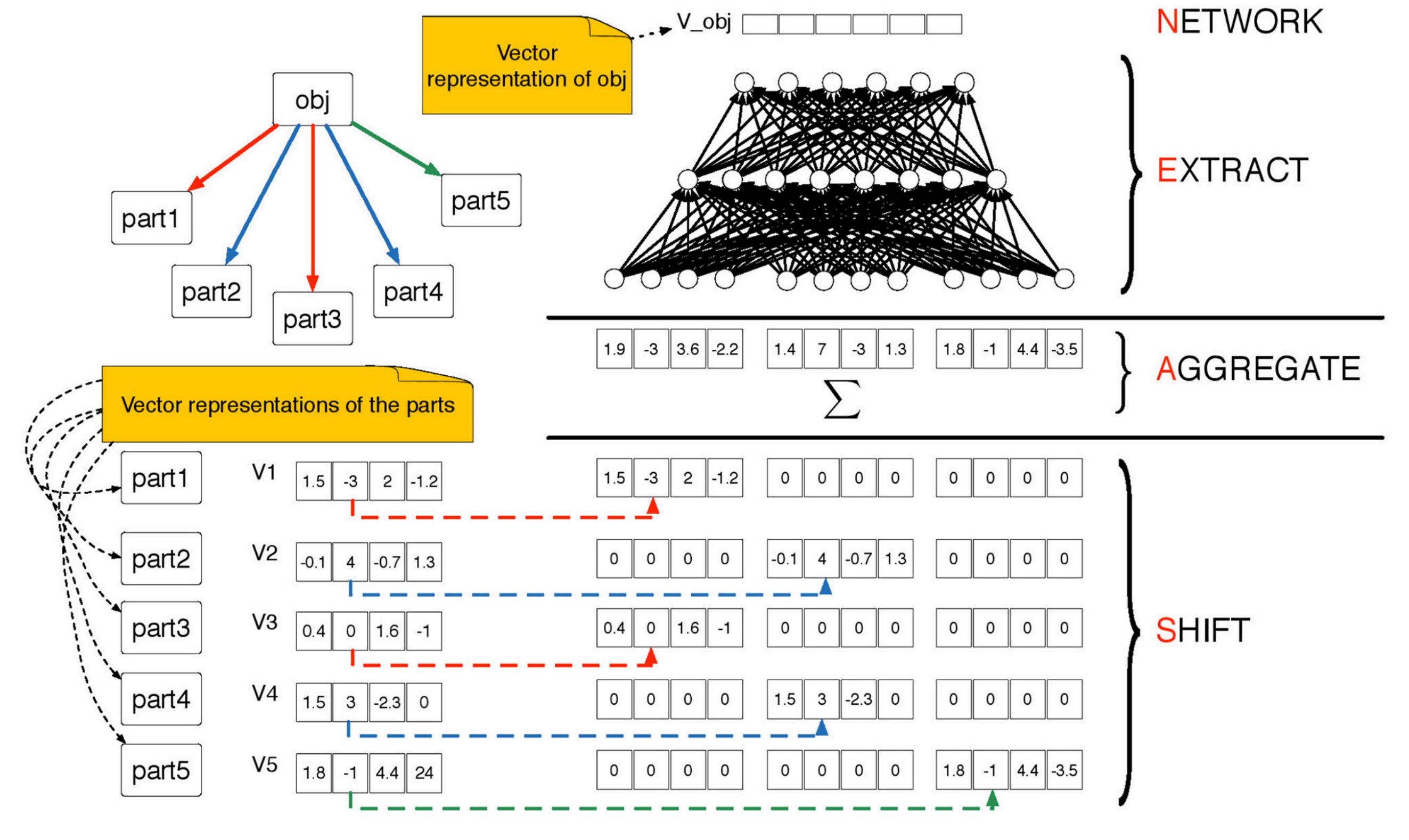

**FIGURE 4 |** Pictorial representation of the SAEN computation explained in Equations 1 and 2. The SAEN computation is unfolded over all the levels of an $\mathcal{H}$-decomposition. On the top-right part we show an object $obj \in S_l$ decomposed into its parts $\{part_i\}_{i=1}^{5} \subseteq S_{l-1}$ from the level below. The parametrized "part of" relation $\mathcal{R}_{l,pi}$ is represented by directed arrows, we use colors (red, blue and green) to distinguish among $\pi$-types. In the bottom-left part of the picture we show that each part is associated to a vectorial representation. In the bottom-right part of the picture we show the *shift* step in which the vector representations of the parts are shifted using the Kronecker product in Equation 1. Then the shifted representation are summed in the aggregation step and in the extract step a feedforward neural is applied in order to obtain the vector representation of object *obj*.

The vector $\mathbf{a}_l(s)$ is obtained in two steps: first, previous level representations $\mathbf{h}_{l-1}(s')$ are *shifted* via the Kronecker product $\otimes$ using an indicator vector $\mathbf{z}_\pi \in \mathbb{R}^{n(l)}$. This takes into account of the membership types $\pi$. Second, shifted representations are *aggregated* with a sum. Note that all representation sizes $d(l)$, $l > 0$ are hyper-parameters that need to be chosen or adjusted.

The shift and aggregate steps are identical to those used in kernel design when computing the explicit feature of a kernel $k(x,z)$ derived from a sum $\sum_{\pi\in\Pi} k_\pi(x,z)$ of base kernels $k_\pi(x,z)$, $\pi \in \Pi$. In principle, it would be indeed possible to turn SAEN into a kernel method by removing the extraction step and define the explicit feature for a kernel on $\mathcal{H}$-decompositions. Removing the extraction step from Equation (1) results in:

$$\mathbf{a}_l(s) = \begin{cases} \mathbf{x}(s) & \text{if } l = 0 \\ \sum\limits_{\pi\in\Pi_l} \sum\limits_{s'\in\mathcal{R}^{-1}_{l,\pi}(s)} \mathbf{z}_\pi \otimes \mathbf{a}_{l-1}(s') & \text{if } l > 0 \end{cases} \quad (3)$$

However, that approach would increase the dimensionality of the feature space by a multiplicative factor $n(l)$ for each level $l$ of the $\mathcal{H}$-decomposition, thus leading to an exponential number of features. When the number of features is exponential, their explicit enumeration is impractical. A possible solution would be to directly define the kernel similarity and keep the features implicit. However, this solution would have space complexity that is quadratic in the number of graphs in the dataset.

When using SAEN, the feature space growth is prevented by exploiting a distributed representation (via a multilayered neural network) during the extraction step. As a result, SAEN can easily cope with $\mathcal{H}$-decompositions consisting of multiple levels.

## 5. EXPLOITING SYMMETRIES FOR DOMAIN COMPRESSION

In this section, we propose a technique, called *domain compression*, which allows us to save memory and speed up the SAEN computation. Domain compression exploits symmetries in $\mathcal{H}$-decompositions to compress them without information loss. This technique requires that the attributes $\mathbf{x}(s)$ of the elements $s$ in the bottom level $S_0$ are categorical.

*Definition 3.* Two objects $a, b$ in a level $S_l$ are *collapsible*, denoted $a \sim b$, if they share the same representation, i.e., $\mathbf{h}_l(a) = \mathbf{h}_l(b)$ for all the possible values of the parameters $\theta_0, \ldots, \theta_l$.

According to *Definition 3*, objects in the bottom level $S_0$ are collapsible when their attributes are identical, while objects at any level $\{S_l\}_{l=1}^{L}$ are collapsible if they are made of the same sets of parts for all the membership types $\pi$.

A compressed level $S_l^{comp}$ is the quotient set of level $S_l$ with respect to the collapsibility relation $\sim$.

Before providing a mathematical formulation of domain compression we provide two examples: in Example 1 we explain the intuition beyond domain compression showing in **Figure 5** the steps that need to be taken to compress a $\mathcal{H}$-decomposition, in Example 2 we provide a pictorial representation of the $\mathcal{H}$-decomposition of a real world graph and its compressed version.

*Example* 1. **Figure 5A** shows the pictorial representation of an $\mathcal{H}$-decomposition whose levels are denoted with the letters of the alphabet A, B, C, D. We name each object using consecutive integers prefixed with the name of the level. We use purple and orange circles to denote the categorical attributes of the objects of the bottom stratum. Directed arrows denote the "part of" relations whose membership type is distinguished using the colors blue and red.

**Figure 5B** shows the domain compression of the $\mathcal{H}$-decomposition in **Figure 5A**. When objects are collapsed the directed arcs coming from their parents are also collapsed. Collapsed arcs are labeled with their cardinality.

**Figures 5C–F** describe the domain compression steps starting from level A until level D.

- In **Figure 5C**, `A3` and `A4` have the same categorical attribute of `A1` (i.e., purple) and they are therefore grouped and collapsed to `A1`. Furthermore, the arrows in the fan-in of `A3` and `A4` are attached to `A1` with the consequent cardinality increase of the red arrows that come from `B3` and `B4`.
- In **Figure 5D** we show the second iteration of domain compression in which objects made of the same parts with the same membership types are collapsed. Both `B1` and `B2` in **Figure 5C** were connected to `A1` with a blue arrow and to `A2` with a red arrow and so they are collapsed. In the same way, `B3` and `B4` are collapsed because in **Figure 5C** they were connected to `A1` with a red arrow with cardinality 2.
- In **Figure 5E** `C1` and `C3` are collapsed because in **Figure 5D** they were both connected to `B1` with a blue arrow and `B3` with a red arrow.
- Finally in **Figure 5F** since `C1` and `C3` were collapsed in the previous step, we increase to 2 the cardinality of the red arrow that connects `D1` and `C1` and remove the red arrow from `D1` to `C3` since `C3` was collapsed to `C1` in **Figure 5E**.

The final result of domain compression is illustrated in **Figure 5B**.

*Example* 2. In **Figure 6** we provide a pictorial representation of the domain compression of an $\mathcal{H}$-decomposition (EGD, described in section 3.1). On the left we show the $\mathcal{H}$-decomposition of a graph taken from the *IMDB-BINARY* dataset (see section 6.1) together with its compressed version on the right.

In order to compress $\mathcal{H}$-decompositions we adapt the lifted linear programming technique proposed by Mladenov et al. (2012) to the SAEN architecture. A matrix $M \in \mathbb{R}^{n\times p}$ with $m \leq n$ distinct rows can be decomposed as the product $DM^{comp}$ where $M^{comp}$ is a compressed version of $M$ in which the distinct rows of $M$ appear exactly once.

*Definition 4.* The Boolean decompression matrix, $D$, encodes the collapsibility relation among the rows of $M$ so that $D_{ij} = 1$ iff the

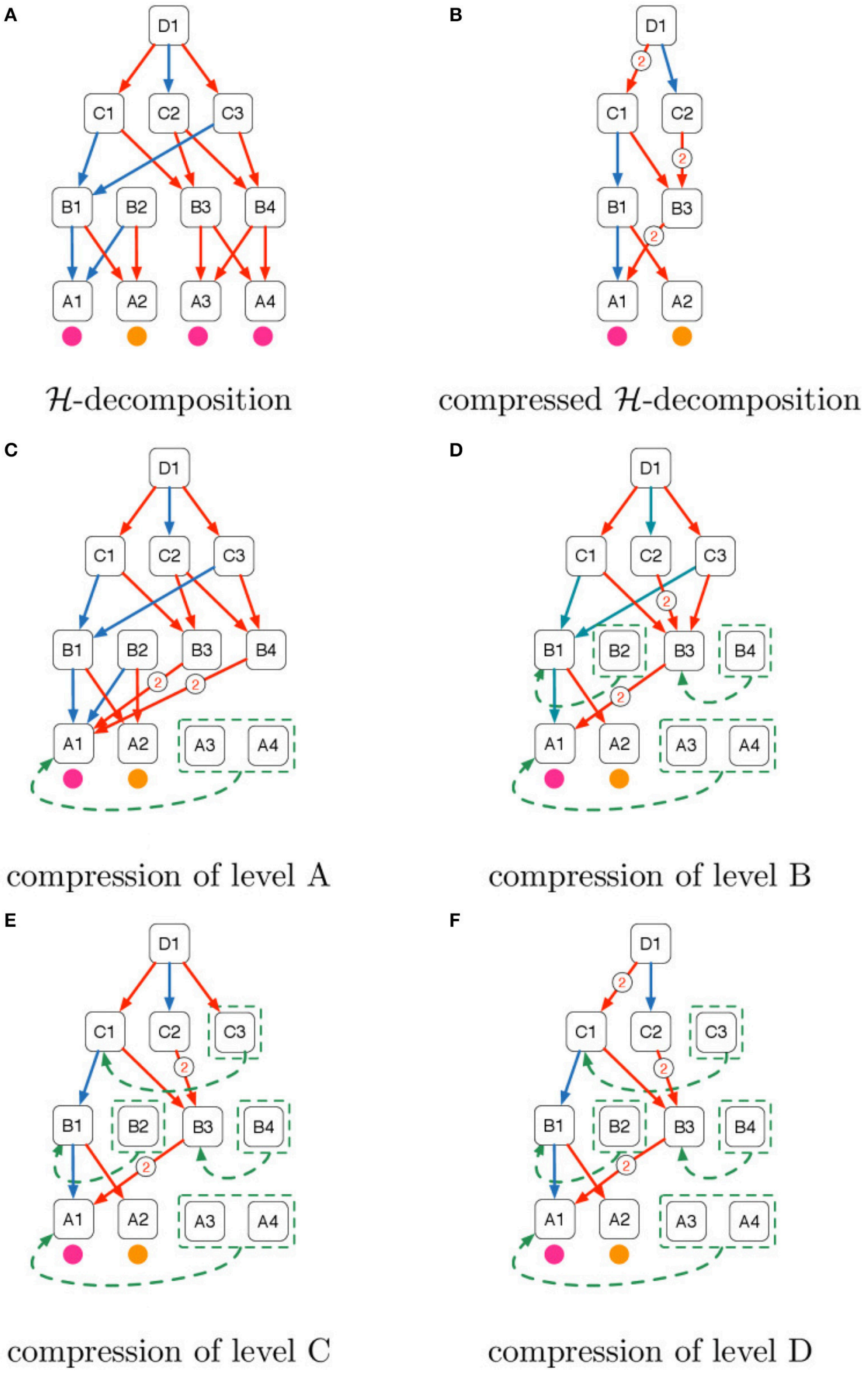

**FIGURE 5 |** Intuition of the domain compression algorithm explained in Example 1. **(A)** $\mathcal{H}$-decomposition, **(B)** compressed $\mathcal{H}$-decomposition, **(C)** compression of level A, **(D)** compression of level B, **(E)** compression of level C, **(F)** compression of level D.

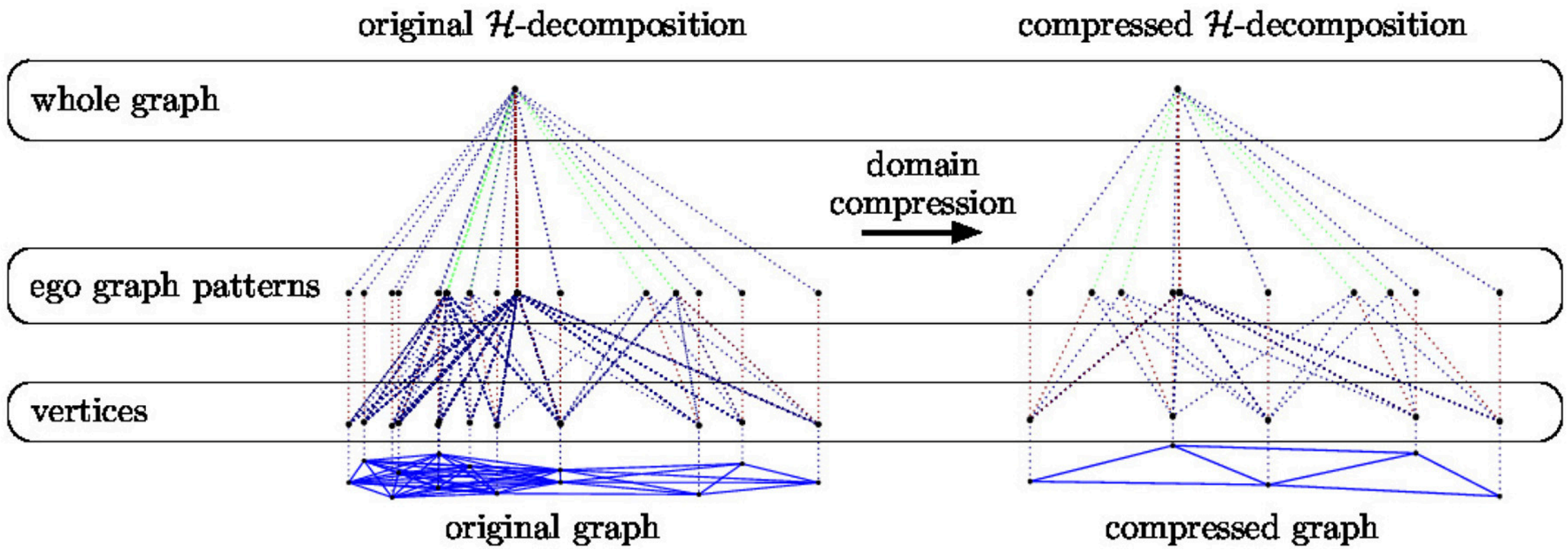

**FIGURE 6 |** Pictorial representation of the $\mathcal{H}$-decomposition of a graph taken from the IMDB - BINARY dataset (see section 6.1) together with its compressed version.

*ith* row of $M$ falls in the equivalence class $j$ of $\sim$, where $\sim$ is the equivalence relation introduced in *Definition 3*[1].

*Example* 3. (Example 1 continued)

The bottom level of the $\mathcal{H}$-decomposition in **Figure 5A** has 4 objects `A1`, `A2`, `A3`, and `A4` with categorical attributes indicated with colors.

Objects `A1`, `A2`, `A4` have a purple categorical attribute while `A3` has a orange categorical attribute. If we give to purple the encoding $[0, 3]$ and to orange the encoding $[4, 1]$ we obtain an attribute matrix

$$X = \begin{bmatrix} 0 & 3 \\ 0 & 3 \\ 4 & 1 \\ 0 & 3 \end{bmatrix} \tag{4}$$

in which each row contains the encoding of the categorical attribute of an object of the bottom stratum and objects were taken with the order `A1`, `A2`, `A3`, `A4`.

Since the rows associated to `A1`, `A3`, `A4` are identical we can compress matrix $X$ to matrix

$$X^{comp} = \begin{bmatrix} 0 & 3 \\ 4 & 1 \end{bmatrix} \tag{5}$$

as we can notice this is the attribute matrix of the compressed $\mathcal{H}$-decomposition shown in **Figure 5B**.

Matrix $X$ can be expressed as the matrix product $DX^{comp}$ between the decompression matrix $D$ and the compressed version of $X^{comp}$ where

$$D = \begin{bmatrix} 1 & 0 \\ 1 & 0 \\ 0 & 1 \\ 1 & 0 \end{bmatrix} \tag{6}$$

and was obtained applying *Definition 4*.

[1] Mladenov et al. (2012) lifts linear programming and defines the equivalence relation induced from the labels obtained by performing color passing on a Gaussian random field. We use an the equivalence relation in *Definition 3* because we are working with $\mathcal{H}$-decompositions.

As explained in Mladenov et al. (2012) a pseudo-inverse $C$ of $D$ can be computed by dividing the rows of $D^\top$ by their sum (where $D^\top$ is the transpose of $D$).

However, it is also possible to compute a pseudo-inverse $C'$ of $D$ by transposing $D$ and choosing one representer for each row of $D^\top$. For each row of $D^\top$ we can simply choose a nonzero element as representer and set all the other to zero.

*Example* 4. The computation of the pseudo-inverse $C$ of the $D$ matrix of Example 3 results in the following equation:

$$C = \begin{bmatrix} 1/3 & 1/3 & 0 & 1/3 \\ 0 & 0 & 1 & 0 \end{bmatrix} \tag{7}$$

the matrix multiplication between the compression matrix $C$ and the $X$ leads to the compressed matrix $X^{comp}$ (i.e., $X^{comp} = CX$).

In the first row of matrix $C$ there are 3 nonzero entries that correspond to the objects `A1`, `A2`, `A4`, while on the second row there is a nonzero entry that corresponds to object `A3`.

As we said above, since we know that the encodings of those objects are identical instead of making the average we could just take a representer.

For example in **Figure 5C** we chose `A1` as representer for `A2` and `A4`, obtaining the compression matrix

$$C' = \begin{bmatrix} 1 & 0 & 0 & 0 \\ 0 & 0 & 1 & 0 \end{bmatrix} \tag{8}$$

In the first row of matrix $C'$ there is a nonzero entry that correspond to the object `A1` (which is the chosen representer), while on the second row there is a nonzero entry that corresponds to object `A3` (as in $C$).

While from the compression point of view we still have $X^{comp} = C'X$, choosing a representer instead of averaging equivalent objects is advantageous when using sparse matrices because the number of nonzero elements decreases.

We apply domain compression to SAEN by rewriting Equations (1, 2) in matrix form.

We rewrite Equation (1) as:

$$A_l = \begin{cases} X & \text{if } l = 0 \\ \mathbf{R}_l \mathbf{H}_{l-1} & \text{if } l > 0 \end{cases} \tag{9}$$

where:

- $A_l \in \mathbb{R}^{|S_l| \times n(l-1)d(l)}$ is the matrix that represents the *shift-aggregated* vector representations of the object of level $S_{l-1}$;
- $X \in \mathbb{R}^{|S_0| \times p}$ is the matrix that represents the $p$-dimensional encodings of the vertex attributes in $V$ (i.e., the rows of $X$ are the $\mathbf{x}_{v_i}$ of Equation 1);
- $\mathbf{R}_l \in \mathbb{R}^{|S_l| \times n(l)|S_{l-1}|}$ is the concatenation

$$\mathbf{R}_l = \left[R_{l,1}, \ldots, R_{l,\pi}, \ldots, R_{l,n(l)}\right] \tag{10}$$

  of the matrices $R_{l,\pi} \in \mathbb{R}^{|S_l| \times |S_{l-1}|}\ \forall \pi \in \Pi_l$ which represent the $\mathcal{R}_{l,\pi}$-convolution relations of Equation (1) whose elements are $(R_{l,\pi})_{ij} = 1$ if $(s', s) \in \mathcal{R}_{l,\pi}$ and 0 otherwise.
- $\mathbf{H}_{l-1} \in \mathbb{R}^{n(l)|S_{l-1}| \times n(l)d(l)}$ is a block-diagonal matrix

$$\mathbf{H}_{l-1} = \begin{bmatrix} H_{l-1} & \ldots & 0 \\ \vdots & \ddots & \vdots \\ 0 & \ldots & H_{l-1} \end{bmatrix} \tag{11}$$

  whose blocks are formed by matrix $H_{l-1} \in \mathbb{R}^{|S_{l-1}| \times d(l)}$ repeated $n(l)$ times. The rows of $H_{l-1}$ are the vector representations $\mathbf{h}_j$ in Equations (1).

Equations (2) is simply rewritten to $H_l = f_l(A_l; \theta_l)$ where $f_l(\cdot; \theta_l)$ is unchanged w.r.t. Equation (2) and is applied to its input matrix $A_l$ row-wise.

**Algorithm 1** DOMAIN-COMPRESSION

DOMAIN-COMPRESSION$(X, R)$

1 $C_0, D_0 = \text{COMPUTE-CD}(X)$
2 $X^{comp} = C_0 X$
3 $R^{comp} = \{\}$
4 **for** $l = 1$ **to** $L$
5 $\quad R^{col_comp} = [R_{l,\pi} D_{l-1},\ \forall \pi = 1, \ldots, n(l)]$
6 $\quad C_l, D_l = \text{COMPUTE-CD}(R^{col_comp})$
7 $\quad$ **for** $\pi = 1$ **to** $n(l)$
8 $\quad\quad R^{comp}_{l,\pi} = C_l R^{col_comp}_{\pi}$
9 **return** $X^{comp}, R^{comp}$

Domain compression in Equation (9) is performed by the DOMAIN-COMPRESSION procedure (see Algorithm 1), which takes as input the attribute matrix $X \in \mathbb{R}^{|S_0| \times p}$ and the part-of matrices $R_{l,\pi}$, and returns their compressed versions $X^{comp}$ and the $R^{comp}_{l,\pi}$, respectively. The algorithm starts by invoking (line 1) the procedure COMPUTE-CD on $X$ to obtain the compression and decompression matrices $C_0$ and $D_0$, respectively. Matrix $C_0$ is used to compress $X$ (line 2). We then iterate over the levels $l = 0, \ldots, L$ of the $\mathcal{H}$-decomposition (line 4) to compress the $R_{l,\pi}$ matrices. Matrices $R_{l,\pi}$ are compressed by right-multiplying them by the decompression matrix $D_{l-1}$ of the previous level $l-1$ (line 5). In this way, we collapse the parts of relation $\mathcal{R}_{l,\pi}$ (i.e., the columns of $R_{l,\pi}$) as these were identified in level $S_{l-1}$ as identical objects (i.e., those objects corresponding to the rows of $X$ or $R_{l-1,\pi}$ collapsed during the previous step). The result is a list $R^{col_comp} = [R_{l,\pi} D_{l-1},\ \forall \pi = 1, \ldots, n(l)]$ of column compressed $R_{l,\pi}$-matrices. We proceed collapsing equivalent objects in level $S_l$, i.e., those made of identical sets of parts: we find symmetries in $R^{col_comp}$ by invoking COMPUTE-CD (line 6) and obtain a new pair $C_l$, $D_l$ of compression, and decompression matrices respectively. Finally, compression matrix $C_l$ is applied to the column-compressed matrices in $R^{col_comp}$ in order to obtain the $\Pi_l$ compressed matrices of level $S_l$ (line 8).

Algorithm 1 allows us to compute the domain compressed version of Equation (9) which can be obtained by replacing $X$ with $X^{comp} = C_0 X$, $R_{l,\pi}$ with $R^{comp}_{l,\pi} = C_l R_{l,\pi} D_{l-1}$, and $H_l$ with $H^{comp}_l$. Willing to recover the original encodings $H_l$ we just need to employ decompression matrix $D_l$ on the compressed encodings $H^{comp}_l$. Indeed, $H_l = D_l H^{comp}_l$. As we can see by replacing $S_l$ with $S^{comp}_l$, the more symmetries are present (i.e., when $|S^{comp}_l| \ll |S_l|$) the greater the domain compression will be.

# 6. EXPERIMENTAL EVALUATION

We perform an experimental evaluation of SAEN on graph classification datasets and answer the following questions:

Q1 How does SAEN compare to the state of the art?
Q2 Can SAEN exploit symmetries in social networks to reduce the memory usage and the runtime?

## 6.1. Datasets

In order to answer the experimental questions we tested our method on six publicly available datasets first proposed by Yanardag and Vishwanathan (2015). These datasets are representative of a wide variety of node degree distributions. While we do not provide a statistical analysis on the node-degree distributions of these datasets, in **Figures 7**, **8** we empirically show the scatter plots of their node-degree frequencies.

- **COLLAB**
  is a dataset where each graph represent the ego-network of a researcher, and the task is to determine the field of study of the researcher between *High Energy Physics*, *Condensed Matter Physics*, and *Astro Physics*.
- **IMDB-BINARY**, **IMDB-MULTI**
  are datasets derived from IMDB where in each graph the vertices represent actors/actresses and the edges connect people which have performed in the same movie. Collaboration graphs are generated from movies belonging to genres *Action* and *Romance* for IMDB-BINARYand *Comedy*, *Romance*, and *Sci-Fi* for IMDB-MULTI, and for each actor/actress in those genres an ego-graph is extracted. The task is to identify the genre from which the ego-graph has been generated.

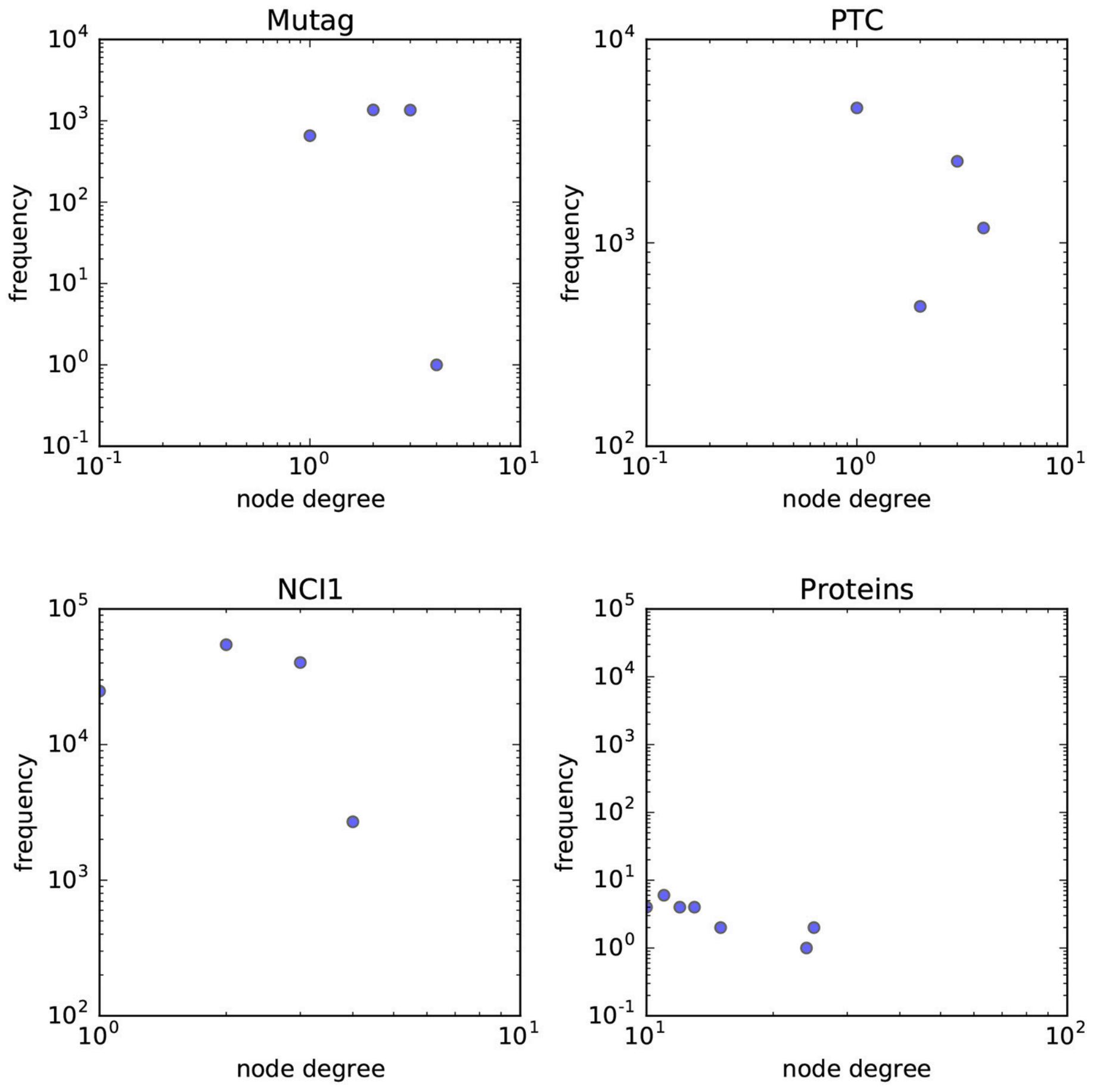

**FIGURE 7 |** Scatterplot of the node degree frequencies in biological datasets visualized in log-log scale.

- **REDDIT-BINARY**, **REDDIT-MULTI5K**, **REDDIT-MULTI12K** are datasets where each graph is derived from a discussion thread from Reddit. In those datasets each vertex represent a distinct user and two users are connected by an edge if one of them has responded to a post of the other in that discussion. The task in *REDDIT-BINARY* is to discriminate between threads originating from a discussion-based subreddit (*TrollXChromosomes*, *atheism*) or from a question/answers-based subreddit (*IAmA*, *AskReddit*). The task in *REDDIT-MULTI5K* and *REDDIT-MULTI12K* is a multiclass classification problem where each graph is labeled with the subreddit where it has originated (*worldnews, videos, AdviceAnimals, aww, mildlyinteresting* for *REDDIT-MULTI5K* and *AskReddit, AdviceAnimals, atheism, aww, IAmA, mildlyinteresting, Showerthoughts, videos, todayilearned, worldnews, TrollXChromosomes* for REDDIT-MULTI12K).

Even if our objective was to build a method suitable for large graphs, for the sake of completeness we also tested our method on some small bioinformatic datasets.

- **MUTAG** Debnath et al. (1991) is a dataset of 188 mutagenic aromatic and heteroaromatic nitro compounds labeled according to whether or not they have a mutagenic effect on the Gramnegative bacterium *Salmonella typhimurium*. **PTC** (Toivonen et al., 2003) is a dataset of 344 chemical compounds that reports the carcinogenicity for male and female rats and it has 19 discrete labels. **NCI1** (Wale et al., 2008) is a dataset of 4,100 examples and is a subset of balanced datasets of chemical compounds screened for ability to suppress or inhibit the growth of a panel of human tumor cell lines. **PROTEINS** (Borgwardt et al., 2005) is a binary classification dataset made of 1,113 proteins. Each protein is represented as a graph where vertices are secondary structure elements (i.e., helices, sheets and turns). Edges connect vertices if they are neighbors in the amino-acid sequence or in the 3D space.

## 6.2. Experiments

### 6.2.1. E1

We experiment with SAEN applying the EGD $\mathcal{H}$-decomposition on *PROTEINS*, *COLLAB*, *IMDB-BINARY*,

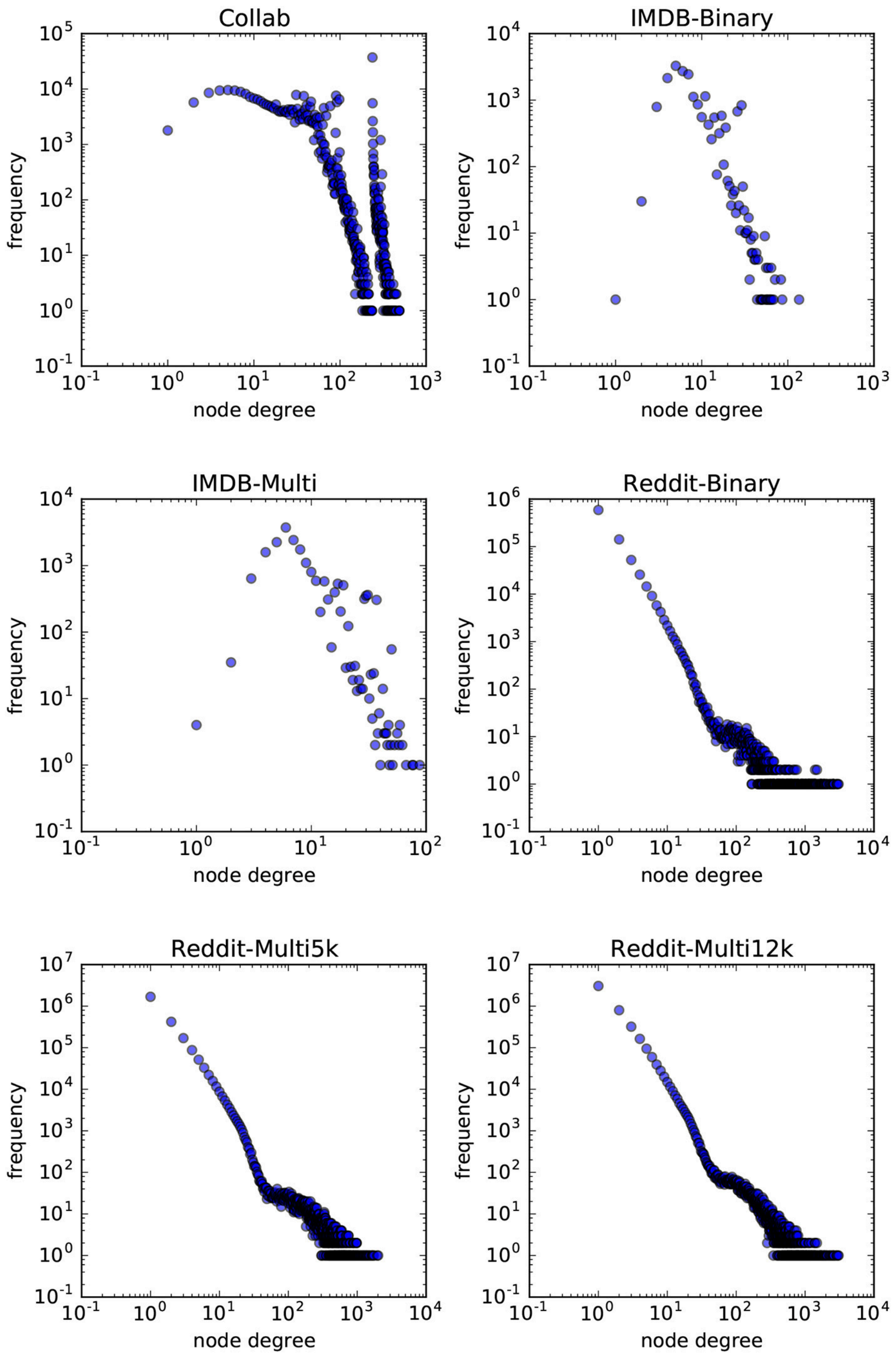

**FIGURE 8 |** Scatterplot of the node degree frequencies in social network datasets visualized in log-log scale.

*IMDB-MULTI*, *REDDIT-BINARY*, *REDDIT-MULTI5K*, and *REDDIT-MULTI12K*, and the NEGD $\mathcal{H}$-decomposition on *MUTAG*, *PTC*, and *NCI1*. We used the colors resulting from 4 iterations of the Weisfeiler-Lehman algorithm (Shervashidze et al., 2011) as identifiers for the ego graphs contained in the bottom level of NEGD.

In order to perform classification we add a cross-entropy loss on the extraction step $h_L(s)$ (see Equation 2) of the top level $L$ (i.e., $L = 2$) of the EGNN $\mathcal{H}$-decomposition. We used Leaky RELUs (Maas et al., 2013) as activation function on all the units of the neural networks $\{f_l(.; \Theta_l)\}_{l=0}^{2}$ of the extraction step (cf. Equation 2).

SAEN was implemented in TensorFlow and in all our experiments we trained the neural network parameters $\{\Theta_l\}_{l=0}^2$ by using the Adam algorithm (Kingma and Ba, 2015) to minimize a cross-entropy loss.

For each dataset, we manually chose the number of layers and units for each level of the part-of decomposition, the coefficient for L2 regularization on the network weights, and the number of training epochs. We ran 10-times 10-fold cross-validation keeping the hyperparameters fixed, measured 10 accuracy values (one for each of the 10 runs of 10-fold cross-validation) and computed mean and standard deviations.

In **Table 2** we provide the results obtained by running our method, Yanardag and Vishwanathan (2015) (DGK), Niepert et al. (2016) (PATCHY-SAN), and Atwood and Towsley (2016) (DCNN) on social network data, while in **Table 5** we provide the results obtained running our method, PATCHY-SAN and DCNN on bioinformatic datasets. For each experiment we provide mean accuracy and standard deviation obtained with the same statistical protocol.

In **Table 3** we report for each dataset the radiuses $r$ of the neighborhood subgraphs used in the EGD decomposition and the number of units in the hidden layers for each level.

#### 6.2.2. E2

In **Table 4** we show the file sizes of the preprocessed datasets before and after the compression together with the data compression ratio[2]. We also estimate the benefit of domain compression from a computational time point of view and report the measurement of the runtime for 10 epochs with and without compression together with the speedup factor.

For the purpose of this experiment, all tests were run on a computer with two 8-cores Intel Xeon E5-2665 processors and 94 GB RAM. SAEN was implemented in Python with the TensorFlow library.

### 6.3. Discussion

#### 6.3.1. A1

As shown in **Table 2**, EGD performs consistently better than the other three methods on all the social network datasets, with the only exception of COLLAB where DCNN outperforms SAEN. This confirms that the chosen $\mathcal{H}$-decomposition is effective on this kind of problems. **Table 1** shows that the average maximum node degree (AMND)[3] of the social network datasets is in the order of $10^2$. SAEN can easily cope with highly-skewed node degree distributions by aggregating distributed representation of patterns while this is not the case for DGK and $PATCHY-SAN$. DGK uses the same patterns of the corresponding non-deep graph kernel used to match common substructures. If the pattern distribution is affected by the degree distribution most of those patterns will not match, making it unlikely for DGK to work well on social network data. $PATCHY-SAN$ employs as patterns neighborhood subgraphs truncated or padded to a size $k$ in order to fit the size of the receptive field of a CNN. However, since Niepert et al. (2016) experiment with $k = 10$, it is not surprising that they perform worst than SAEN on *COLLAB*, *IMDB-MULTI*, *REDDIT-MULTI5K*, and *REDDIT-MULTI12K* since a small $k$ causes the algorithm to throw away most of the subgraph; a more sensible choice for $k$ would have been the AMND of each graph (i.e., 74, 12, 204, and 162 respectively, cf. **Tables 1**, **2**). *DCNN* obtained the best results on COLLAB and was competitive on IMDB-BINARY and IMDB-MULTI. However, this method needs to compute and store the power series $P^2, \ldots, P^H$ where $P$ is the transition matrix of a graph and so it has space complexity $O(|V|^2 H)$. While the memory usage of the algorithm did not lead to out-of-memory errors, we noticed a huge increase of the runtime when dealing with larger graphs. In fact, the algorithm was not able to complete the execution on REDDIT-BINARY, REDDIT-MULTI5K, and REDDIT-MULTI12K in a time budget of 10 days.

**TABLE 1** | Statistics of the datasets used in our experiments.

| Dataset | Size | Avg. vertices | Avg. max. degree |
|---|---|---|---|
| COLLAB | 5, 000 | 74.49 | 73.62 |
| IMDB-BINARY | 1, 000 | 19.77 | 18.77 |
| IMDB-MULTI | 1, 500 | 13.00 | 12.00 |
| REDDIT-BINARY | 2, 000 | 429.62 | 217.35 |
| REDDIT-MULTI5K | 5, 000 | 508.51 | 204.08 |
| REDDIT-MULTI12K | 11, 929 | 391.40 | 161.70 |
| MUTAG | 188 | 17.93 | 3.01 |
| PTC | 344 | 25.56 | 3.73 |
| NCI1 | 4, 110 | 29.87 | 3.34 |
| PROTEINS | 1, 113 | 39.06 | 5.79 |

**Table 5** compares the results of SAEN with the best $PATCHY-SAN$ instance on chemoinformatics and bioinformatics datasets. Results obtained by SAEN are comparable with the ones obtained by Niepert et al. (2016) on *NCI*1 and *PROTEINS*, confirming that SAEN is best suited for graphs with large degrees. Moreover, SAEN does not perform well on *MUTAG* and *PTC*, as these datasets are too small to afford the highly expressive representations that SAEN can learn and in spite of regularization with L2 we consistently observed significant overfitting.

#### 6.3.2. A2

The compression algorithm has proven to be effective in improving the computational cost of our method. Most of the datasets halved their runtimes while maintaining the same expressive power. Moreover, we reduced the memory usage on the largest datasets to less than 40% of what would have been necessary without compression.

## 7. RELATED WORKS

When learning with graph inputs two fundamental design aspects that must be taken into account are: the choice of the

[2]The size of the uncompressed files are shown for the sole purpose of computing the data compression ratio. Indeed the last version of our code compresses the files on the fly.

[3]The AMND for a given dataset is obtained by computing the maximum node degree of each graph and then averaging over all graphs.

**TABLE 2 |** Results on social network datasets (taken from Yanardag and Vishwanathan, 2015; Niepert et al., 2016, for DGK and PATCHY-SAN, respectively, and run using the implementation made available by Atwood and Towsley, 2016, https://github.com/jcatw/dcnn, for DCNN).

| Dataset | DGK | PATCHY-SAN | DCNN | SAEN |
|---|---|---|---|---|
| COLLAB | $73.09 \pm 0.25$ | $72.60 \pm 2.16$ | $79.60 \pm 0.28$ | $78.50 \pm 0.69$ |
| IMDB-BINARY | $66.96 \pm 0.56$ | $71.00 \pm 2.29$ | $70.48 \pm 0.29$ | $71.59 \pm 1.20$ |
| IMDB-MULTI | $44.55 \pm 0.52$ | $45.23 \pm 2.84$ | $47.92 \pm 0.56$ | $48.53 \pm 0.76$ |
| REDDIT-BINARY | $78.04 \pm 0.39$ | $86.30 \pm 1.58$ | – | $87.22 \pm 0.80$ |
| REDDIT-MULTI5K | $41.27 \pm 0.18$ | $49.10 \pm 0.70$ | – | $53.63 \pm 0.51$ |
| REDDIT-MULTI12K | $32.22 \pm 0.10$ | $41.32 \pm 0.42$ | – | $45.27 \pm 0.30$ |

*Missing values indicate that the algorithm did not complete in the given time budget.*

**TABLE 3 |** Parameters used for the *EGD* decompositions for each datasets.

| Dataset | Decomposition | Hidden units | | |
|---|---|---|---|---|
| | | $S_0$ | $S_1$ | $S_2$ |
| COLLAB | EGD, $r = 1$ | $15 - 5$ | $5 - 2$ | $5 - 3$ |
| IMDB-BINARY | EGD, $r = 2$ | 2 | $5 - 2$ | $5 - 3 - 1$ |
| IMDB-MULTI | EGD, $r = 2$ | 2 | $5 - 2$ | $5 - 3$ |
| REDDIT-BINARY | EGD, $r = 1$ | $10 - 5$ | $5 - 2$ | $5 - 3 - 1$ |
| REDDIT-MULTI5K | EGD, $r = 1$ | 10 | 10 | $6 - 5$ |
| REDDIT-MULTI12K | EGD, $r = 1$ | 10 | 10 | $20 - 11$ |
| MUTAG | NEGD | 20 | $40 - 20$ | $40 - 20 - 1$ |
| PTC | NEGD | 50 | $100 - 50$ | $100 - 50 - 1$ |
| NCI1 | NEGD | 50 | $100 - 50$ | $100 - 50 - 1$ |
| PROTEINS | EGD, $r = 3$ | 3 | 3 | $9 - 6 - 1$ |

**TABLE 4 |** Comparison of sizes and runtimes (for 10 epochs) of the datasets before and after the compression.

| Dataset | Size (mb) | | | Runtime | | |
|---|---|---|---|---|---|---|
| | Original | Comp. | Ratio | Original | Comp. | Speedup |
| *COLLAB* | 337 | 119 | 0.35 | $2' 27''$ | $1' 06''$ | 2.23 |
| *IMDB-BINARY* | 24 | 18 | 0.75 | $8''$ | $6''$ | 1.33 |
| *IMDB-MULTI* | 31 | 25 | 0.81 | $19''$ | $17''$ | 1.12 |
| *REDDIT-BINARY* | 129 | 47 | 0.36 | $47''$ | $16''$ | 2.94 |
| *REDDIT-MULTI5K* | 368 | 132 | 0.36 | $2' 10''$ | $55''$ | 2.36 |
| *REDDIT-MULTI12K* | 712 | 287 | 0.40 | $4' 25''$ | $2' 02''$ | 2.17 |

**TABLE 5 |** Comparison of accuracy on bio-informatics datasets (taken from Niepert et al., 2016 for PATCHY-SAN, and run using the implementation made available by Atwood and Towsley, 2016, https://github.com/jcatw/dcnn, for DCNN).

| Dataset | PATCHY-SAN Niepert et al. (2016) | DCNN Atwood and Towsley (2016) | SAEN (our method) |
|---|---|---|---|
| MUTAG | $92.63 \pm 4.21$ | $64.18 \pm 2.97$ | $82.48 \pm 1.43$ |
| PTC | $62.29 \pm 5.68$ | $55.81 \pm 0.00$ | $56.80 \pm 1.40$ |
| NCI1 | $78.59 \pm 1.89$ | $60.57 \pm 0.35$ | $78.62 \pm 0.40$ |
| PROTEINS | $75.89 \pm 2.76$ | $66.06 \pm 0.57$ | $72.73 \pm 0.96$ |

pattern generator and the choice of the matching operator. The former decomposes the graph input in substructures while the latter allows to compare the substructures.

Among the patterns considered from the graph kernel literature we have paths, shortest paths, walks (Kashima et al., 2003), subtrees (Ramon and Gärtner, 2003; Shervashidze et al., 2011) and neighborhood subgraphs (Costa and De Grave, 2010). The similarity between graphs $G$ and $G'$ is computed by counting the number of matches between their common substructures (i.e., a kernel on the sets of the substructures). The match between two substructures can be defined by using graph isomorphism or some other weaker graph invariant. One advantage of graph kernels such as the Weisfeiler-Lehman subtree kernel (*WLST*) (Shervashidze et al., 2011) and the Neighborhood Subgraph Pairwise Distance Kernel (NSPDK) (Costa and De Grave, 2010) is the possibility to efficiently compute explicit feature vectors, thus avoiding to solve the optimization problem in the dual. As we explained in section 4, we could in principle turn SAEN into a graph kernel by removing the extraction step; this approach however would be impractical because of the exponential growth of the number of features. Additionally, the corresponding feature map would be fixed before observing data, as it happens with all graph kernels. SAEN, like other neural network models, can learn graph representations.

Micheli (2009) proposed neural networks for graphs (NN4G), a feedforward neural network architecture for $\mathbf{l}$-attributed graphs that first applies a single layer neural network to the vertex attributes $\mathbf{l}(v)$ to produce the an initial encoding $x_1(v)$ for the vertices $v$ in the graph $G$ and then iteratively find new vector representations $x_i(v)$ for the vertices of the input graph $G$. During the successive iterations the state encoding $x_i(v)$ of a vertex $v$ is obtained by stacking a single neural network layer with sigmoid activation functions that take as input the continuous attributes $\mathbf{l}(v)$ of $v$ and the state encodings $x_{i'}(u)$ of the neighbors $u$ of $v$ during all the previous iterations $i' < i$. Finally, NN4G can either learn an output representation $y_o(p)$ for the vertices (i.e., $p = v$) or for the whole graph (i.e., $p = G$). While the former is obtained by stacking a single layer neural network over the encoding of the vertices produced across all the iterations, the latter is obtained by aggregating for each iteration $i$ the vertex representations $x_i(v)$ over the vertices $v$ of $G$, producing a graph representation $X_i(G)$ for each iteration $i$ and then stacking stacking a single layer neural network. Differently from RNNs, both SAEN and NN4G can learn from graph inputs without imposing weight sharing and using feedforward neural networks. However, while in both NN4G and RNNs the computation is bound to follow the connectivity of the input graph, SAEN has a computation model that follows the connectivity of $\mathcal{H}$-decompositions which can be specified by the user. Moreover, the SAEN user can specify how the vector encoding should be shifted before the aggregation by using the $\pi$-membership types of the $\mathcal{H}$-decompositions. Furthermore, SAEN can be trained end-to-end with backpropagation while NN4G was not. Indeed, at each iteration of the computation of a state encoding NN4G *freezes* the weights of the previous iterations.

Deep graph kernels (DGK) (Yanardag and Vishwanathan, 2015) upgrade existing graph kernels with a feature reweighing schema. DGKs represent input graphs as a corpus of substructures (e.g., graphlets, Weisfeiler-Lehman subtrees, vertex pairs with shortest path distance) and then train vector embeddings of substructures with CBOW/Skip-gram models[4]. Each graph-kernel feature (i.e., the number of occurrences of a substructure) is reweighed by the 2-norm of the vector embedding of the corresponding substructure. Experimental evidence shows that DGKs alleviate the problem of diagonal dominance in GKs. However, DGKs inherit from GKs a flat representation (i.e., just one layer of depth) of the input graphs and the vector representations of the substructures are not trained end-to-end as SAEN would do.

The use of CNNs for graphs has been initially proposed by Bruna et al. (2014) and subsequently improved by Defferrard et al. (2016). These works extend convolutions from signals defined on time or regular grids to domains defined by arbitrary undirected graphs. These methods can not be directly applied to the graph classification problem where each graph in the dataset has a different size and structure. $PATCHY-SAN$ (Niepert et al., 2016) is able to apply CNNs to the graph classification problem by decomposing graphs into a fixed number of neighborhood subgraphs and casting them to fixed-size receptive fields. Both steps involve either padding or truncation in order to meet the fixed-size requirements. The truncation operation can be detrimental for the statistical performance of the downstream CNN since it throws away part of the input graph. On the other hand SAEN is able to handle structured inputs of variable sizes without throwing away part of the them. A related neural network architecture was recently introduced by Tibo et al. (2017) to extend the multi-instance learning framework to data represented as bags of bags of instances. That network can be seen as a special case of SAEN using maximum as the aggregation operator and no $\pi$-types (i.e., no shifts).

Atwood and Towsley (2016) proposed a diffusion-convolutional neural network (DCNN) that can be used for both whole-graph and node classification. A transition matrix $P$ (derived by normalizing the graph adjacency matrix) is used to propagate learned representations of vertices for $H$ iterations. In the node classification setting a neural network is applied on the nodes representations, while in the graph classification setting a neural network is applied on the aggregation of the node representations. While this method can be applied to both node and graph classification, it has some scalability issues. Indeed it requires to store the power series $P, P^2, \ldots, P^H$ and this operation has $O(|V|^2H)$ space complexity. For this reason this method can lead to out-of-memory errors when dealing with large graphs.

[4] The CBOW/Skip-gram models receive as inputs cooccurrences among substructures sampled from the input graphs.

Several other researchers have studied methods for computing node features and thus solving the node classification problem over graphs. These methods have not been applied to the whole-graph classification problem. On the other hand, SAEN, as described in this paper, cannot be directly applied to the node classification problem.

One approach to learn node representations was introduced in Kipf and Welling (2017) within the semi-supervised learning setting using convolutional networks. *GRAPHSAGE* (Hamilton et al., 2017a) generates representations for vertices of a graph using an algorithm inspired by the Weisfeiler-Lehman isomorphism test. The initial representation $\mathbf{h}_v^0$ of each node $v$ is set to the corresponding attribute vector $\mathbf{x}_v$. Then, for a fixed number of times $K$, a new representation for $v$ is built by applying a single neural network layer to the concatenation of the node's previous representation $\mathbf{h}_v^{k-1}$ and an aggregated representation $\mathbf{h}_{\mathcal{N}(v)}^k$ of the neighborhood of $v$ (according to a neighborhood function $\mathcal{N}(v)$). The approach used by *GRAPHSAGE* to propagate representations is similar to the application of SAEN's shift-aggregate operators between level 0 and 1 of ego graph decompositions; unlike SAEN, however, the new node descriptor is built via a single neural network layer instead of a generic extract operation. Furthermore, the algorithm in *GRAPHSAGE* is forced to use a fixed neighborhood function for all the propagation steps, whereas SAEN is explicitly designed to be able to handle different "part of" relationships at different levels of the hierarchy. Finally, while the special handling of the neighborhood's center is hardcoded in *GRAPHSAGE*, in SAEN the more generic $\pi$-types mechanism is used to describe the role of each node in the ego graphs, and of each ego graph in the whole graph.

Hamilton et al. (2017b) proposed a comprehensive review of methods to embed vertices and graphs. Sum-based approaches such as the ones proposed by Dai et al. (2016) and Duvenaud et al. (2015) build graph representations by summing node embeddings or edge embeddings; these approaches however cannot represent more complex decompositions and cannot distinguish between vertices with different roles.

The exploitation of symmetries that we have proposed in section 5 for compressing relational structures is related to some algorithmic ideas that have been previously proposed for lifted inference in graphical models.

In particular, counting belief propagation (CBP) (Kersting et al., 2009) exploits symmetries in factor graphs in order to speed up belief propagation. Our goal is instead to improve space and time requirements for the SAEN computation.

In CBP, nodes and factors that *send the same messages* are grouped into clusternodes and clusterfactors, respectively, leading to a compressed factor graph. In Algorithm 1 we group together objects in the $\mathcal{H}$-decomposition that *produce identical representations* under the computation defined by SAEN.

As noted in (Mladenov et al., 2014), the compressed factor graph approach of CBP finds the same clusternodes and clusterfactors that would be obtained by running the 1-dimensional Weisfeiler-Lehman algorithm on the uncompressed factor graph. Domain compression in Algorithm 1 is also obtained by a special form of message passing but in this case finalized at exchanging results of the intermediate representations computed by SAEN.

## 8. CONCLUSIONS

Hierarchical decompositions introduce a novel notion of depth in the context of learning with structured data, leveraging the nested part-of-parts relation. In this work, we defined a simple architecture based on neural networks for learning representations of these hierarchies. We showed experimentally that the approach is particularly well-suited for dealing with graphs that are large and have high degree, such as those that naturally occur in social network data. Our approach is also effective for learning with smaller graphs, such as those occurring in chemoinformatics and bioinformatics, although in these cases the performance of SAEN does not exceed the state-of-the-art established by other methods. A second contribution of this work is the domain compression algorithm, which greatly reduces memory usage and allowed us to halve the training time on the largest datasets.

## AUTHOR CONTRIBUTIONS

FO designed the initial version of SAEN and domain compression algorithms, generated the first draft of the manuscript and contributed to edits and updates of the manuscript. DB performed the experiments and contributed to edits and updates of the manuscript. PF supervised the work and contributed to edits and updates of the manuscript.

## FUNDING

We gratefully acknowledge an NVIDIA hardware grant.

**Conflict of Interest Statement:** The authors declare that the research was conducted in the absence of any commercial or financial relationships that could be construed as a potential conflict of interest.